\documentclass[sigconf,natbib=true]{acmart}
\makeatletter
\def\@ACM@checkaffil{
    \if@ACM@instpresent\else
    \ClassWarningNoLine{\@classname}{No institution present for an affiliation}%
    \fi
    \if@ACM@citypresent\else
    \ClassWarningNoLine{\@classname}{No city present for an affiliation}%
    \fi
    \if@ACM@countrypresent\else
        \ClassWarningNoLine{\@classname}{No country present for an affiliation}%
    \fi
}
\makeatother
\usepackage{algorithm}
\usepackage{microtype}
\usepackage{algcompatible,amsmath}
\usepackage{graphicx}
\usepackage{url}
\usepackage{multirow}
\usepackage{comment}
\usepackage{pifont}
\usepackage{balance}
\usepackage{mathtools}
\usepackage{xcolor}
\usepackage{colortbl}
\usepackage{bbm}
\usepackage{enumitem}
\usepackage{marvosym}
\usepackage{textcomp}
\usepackage{array}
\usepackage{booktabs}
\usepackage{tabularx}
\usepackage{CJKutf8}
\usepackage{algorithm}  
\usepackage{algorithmicx}  
\usepackage{algpseudocode} 

\usepackage{fancyhdr}
\usepackage{newfloat}
\usepackage{listings}
\floatstyle{ruled}
\newfloat{listing}{tb}{lst}{}
\floatname{listing}{Listing}

\newcommand{\squishlist}{
\begin{list}{$\bullet$}
{ \usecounter{Lcount}
\setlength{\itemsep}{0pt}
\setlength{\parsep}{0pt}
\setlength{\topsep}{0pt}
\setlength{\partopsep}{0pt}
\setlength{\leftmargin}{2em}
\setlength{\labelwidth}{1.5em}
\setlength{\labelsep}{0.5em} } }
\newcommand{\squishend}{
\end{list} }
\newcommand*\circled[1]{\kern-2.5em%
  \put(0,4){\color{white}\circle*{18}}\put(0,4){\circle{10}}%
  \put(-3,0){\color{black}\bfseries#1}~~}
\usepackage{tikz} 

\usepackage{enumitem}
\newcommand{\modelname}{{\textsc{\textsf{ReRead}}}}
\makeatletter
\newcommand{\thickhline}{%
    \noalign {\ifnum 0=`}\fi \hrule height 2pt
    \futurelet \reserved@a \@xhline
}

\definecolor{codegreen}{rgb}{0,0.6,0}
\definecolor{codegray}{rgb}{0.5,0.5,0.5}
\definecolor{codepurple}{rgb}{0.58,0,0.82}
\definecolor{backcolour}{rgb}{0.95,0.95,0.92}

\lstdefinestyle{mystyle}{
    backgroundcolor=\color{backcolour},   
    commentstyle=\color{codegreen},
    keywordstyle=\color{magenta},
    numberstyle=\tiny\color{codegray},
    stringstyle=\color{codepurple},
    basicstyle=\ttfamily\footnotesize,
    breakatwhitespace=false,         
    breaklines=true,                 
    captionpos=b,                    
    keepspaces=true,                 
    numbers=left,                    
    numbersep=5pt,                  
    showspaces=false,                
    showstringspaces=false,
    showtabs=false,                  
    tabsize=2
}

\copyrightyear{2023}
\acmYear{2023}
\setcopyright{rightsretained}
\acmConference[SIGIR '23]{Proceedings of the 46th International ACM SIGIR Conference on Research and Development in Information Retrieval}{July 23--27, 2023}{Taipei, Taiwan.}
\acmBooktitle{Proceedings of the 46th International ACM SIGIR Conference on Research and Development in Information Retrieval (SIGIR '23), July 23--27, 2023, Taipei, Taiwan}\acmDOI{10.1145/3539618.3592049}
\acmISBN{978-1-4503-9408-6/23/07}

\usepackage{etoolbox}
\makeatletter
\patchcmd{\maketitle}{\@copyrightpermission}{
   \begin{minipage}{0.3\columnwidth}
     \href{http://creativecommons.org/licenses/by/4.0/}{\includegraphics[width=0.90\textwidth]{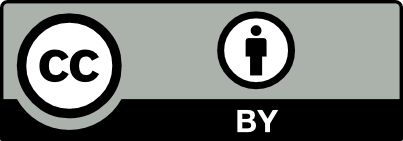}}
   \end{minipage}\hfill
   \begin{minipage}{0.7\columnwidth}
     \href{http://creativecommons.org/licenses/by/4.0/}{This work is licensed under a Creative Commons Attribution International 4.0 License.}
   \end{minipage}
}{}{}

\makeatother
\begin{document}

\title{Read it Twice: Towards Faithfully Interpretable Fact Verification by Revisiting Evidence}

\author{Xuming Hu}
\affiliation{%
  \institution{Tsinghua University}
}
\email{hxm19@mails.tsinghua.edu.cn}

\author{Zhaochen Hong}
\affiliation{%
  \institution{Tsinghua University}
}
\email{hongzc20@mails.tsinghua.edu.cn}

\author{Zhijiang Guo}
\affiliation{%
  \institution{University of Cambridge}
  }
\email{zg283@cam.ac.uk}

\author{Lijie Wen}
\affiliation{%
  \institution{Tsinghua University}
  }
\email{wenlj@tsinghua.edu.cn}

\author{Philip S. Yu}
\affiliation{%
 \institution{University of Illinois at Chicago}
 }
\email{psyu@cs.uic.edu}
\renewcommand{\shortauthors}{Hu, et al.}


\begin{abstract}
Real-world fact verification task aims to verify the factuality of a claim by retrieving evidence from the source document. The quality of the retrieved evidence plays an important role in claim verification. Ideally, the retrieved evidence should be \textit{faithful} (reflecting the model's decision-making process in claim verification) and \textit{plausible} (convincing to humans), and can improve the \textit{accuracy} of verification task. Although existing approaches leverage the similarity measure of semantic or surface form between claims and documents to retrieve evidence, they all rely on certain heuristics that prevent them from satisfying all three requirements. In light of this, we propose a fact verification model named {\modelname} to retrieve evidence and verify claim that: (1) Train the evidence retriever to obtain interpretable evidence (i.e., faithfulness and plausibility criteria); (2) Train the claim verifier to revisit the evidence retrieved by the optimized evidence retriever to improve the accuracy.
The proposed system is able to achieve significant improvements upon best-reported models under different settings. 
\end{abstract}







\maketitle
\section{Introduction}
\label{intro}
\begin{figure}
    \centering
    \includegraphics[scale=0.31]{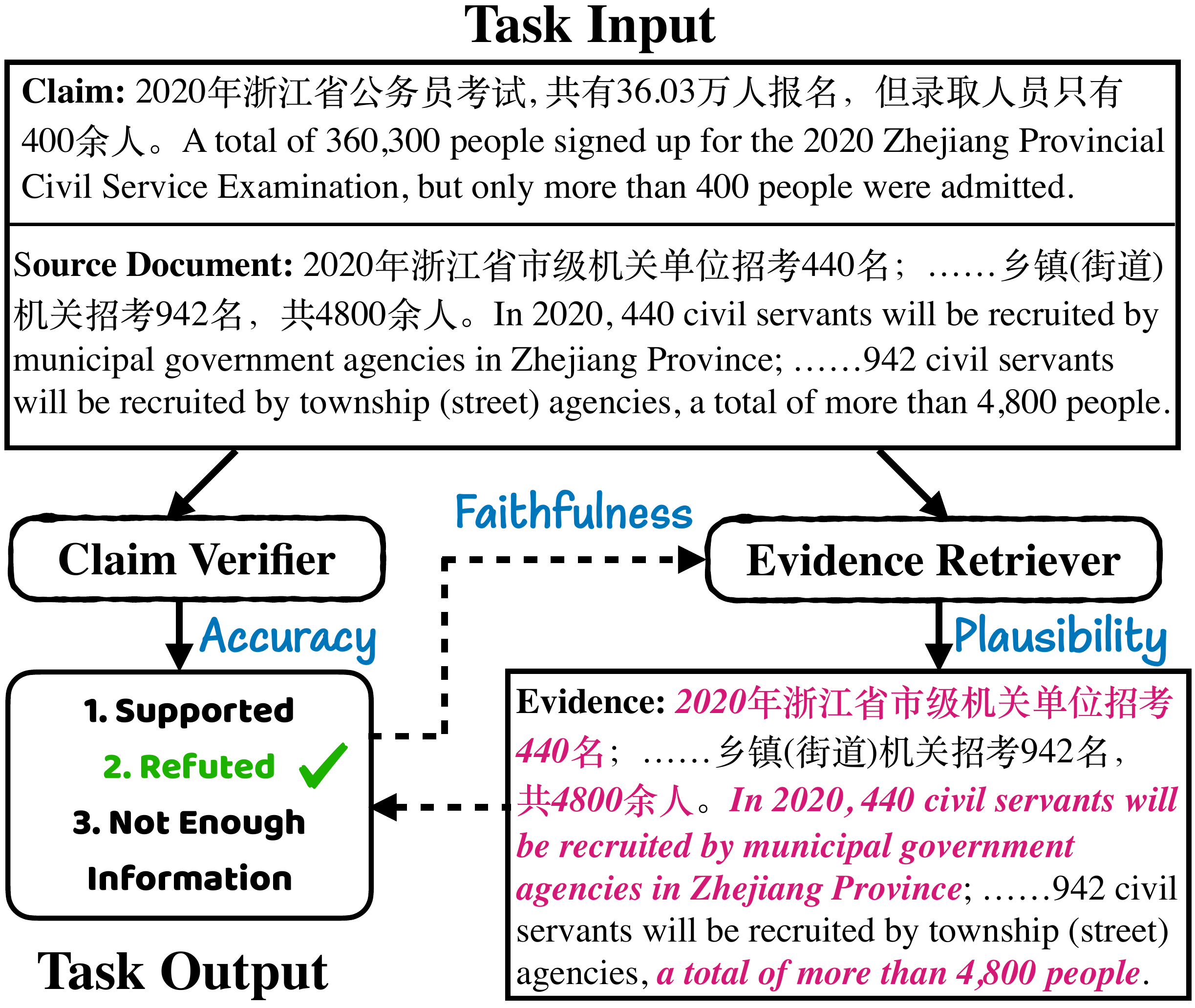}
    \vspace{-4mm}
    \caption{A case of {\modelname}. The evidence retriever should retrieved evidence which could give the plausible reason why the verification result is ``Refuted'' and reflect the verifier's decision-making process. With the training of the evidence retriever, it can provide the verifier with better evidence to revisit and improve the accuracy of the fact verification task.}
    \label{fig:example}
    \vspace{-5mm}
\end{figure}

The spread of misinformation has become a significant issue in today's society, particularly in the digital age where information can be easily disseminated and shared across various platforms \cite{vosoughi2018spread,botnevik2020brenda,pradeep2021vera}. As such, fact verification has emerged as a crucial task in combating this issue by assessing the factuality of claims made in written or spoken language \cite{Adair2017ProgressT,vo2018rise,2018graves,chen2022gere,nielsen2022mumin}. To achieve this goal, it is essential to have appropriate evidence that supports or refutes a claim. Therefore, how to retrieve suitable evidence from a large number of source documents is a key component of fact verification.

As shown in Figure \ref{fig:example}, a real-world claim from Chinese social media and corresponding source document are retrieved through Google search engine. We need to retrieve \textit{faithful} (reflecting the decision-making process of the verifier in claim verification) and \textit{plausible} (explaining the reason for the factuality of the claim) evidence from the noisy document to improve the task \textit{accuracy} of claim verification \cite{zeng2021automated,guo2022survey}. In this case, evidence such as ``more than 4800 people'' needs to be retrieved to counter the claim of ``only more than 400 people''. Although evidence plays a crucial role in fact verification, early automated fact verification attempts disregarded this, and solely relied on the surface patterns of the claim to verify it while ignoring the information that evidence provides \cite{Rashkin2017TruthOV,turenne2018rumour}. Consequently, these approaches were unable to identify well-camouflaged misinformation \cite{Schuster2020TheLO}. Recent efforts to address this issue involve asking annotators to create claims and evidence by mutating sentences from Wikipedia articles \cite{Thorne2018FEVERAL,aly1feverous}. However, these synthetic claims generated from Wikipedia cannot serve as a substitute for real-world claims that circulate in the media ecosystem. As a result, other works resorted to scraping claims from fact-checking sites and using search engines to find supporting documents \cite{ye2019multi,gupta2021x,hu2022chef}. However, the source documents retrieved in this way is often noisy, which hinders the accuracy of verification task. To address this, \citet{hu2022chef} retrieve relevant evidence from the source documents by measuring semantic similarity between the claim and the evidence and \citet{gupta2021x} develop an attention-based evidence aggregation model. However, these methods all rely on certain heuristics and cannot simultaneously satisfy the three requirements of being faithful, plausible, and improving the fact verification accuracy.

We propose the novel real-world fact verification model {\modelname}, which meets three key requirements by: (1) Training an evidence retriever for interpretable evidence based on faithfulness and plausibility criteria; (2) Training a claim verifier to re-evaluate evidence from the optimized retriever, enhancing accuracy. As illustrated in Figure \ref{fig:example}, {\modelname} fine-tunes the verifier using labeled data, then utilizes it to help the retriever obtain faithful evidence. The retriever also uses gold evidence to boost plausibility. Improved evidence provided by the trained retriever allows the verifier to refine accuracy. Our main contributions include: (1) A novel model for retrieving faithful and plausible evidence, increasing verification accuracy; (2) Experiments demonstrating a 4.31\% F1 performance gain over the SOTA baseline on a real-world dataset, with extensive analysis validating {\modelname}'s effectiveness.


 \begin{figure}
    \centering
    \includegraphics[scale=0.33]{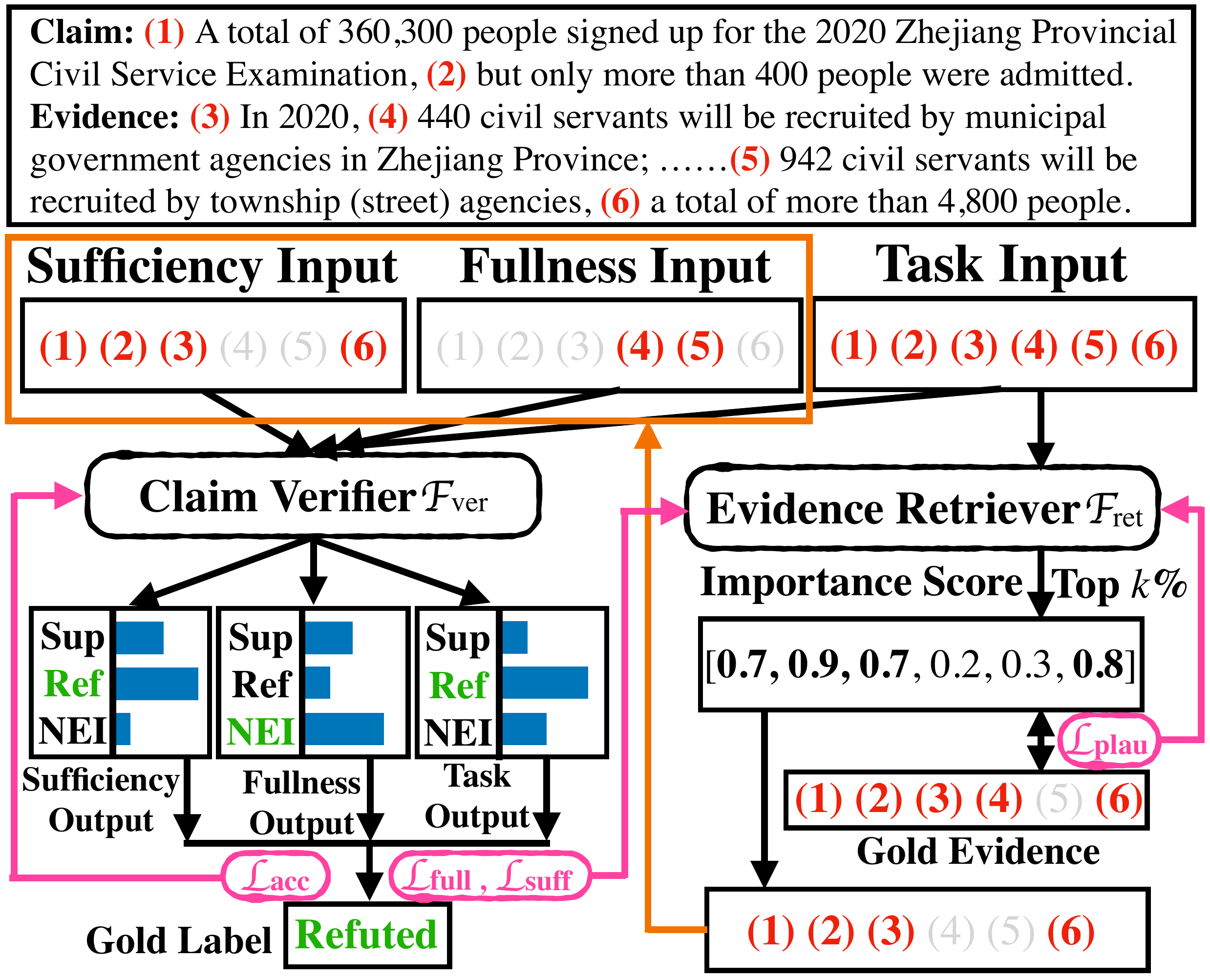}
    \vspace{-3mm}
    \caption{Architecture of {\modelname}. }
    \label{fig:overview}
    \vspace{-5mm}
\end{figure}

\section{Training Goal Analysis}
We have three training goals: (1) The retrieved evidence needs to have \textbf{Faithfulness}, which means how accurately the evidence reflects the true reasoning process of the verifier to predict the verification label \cite{jacovi2020towards}. We use two metrics: \textbf{Fullness} reflects the change in probability of the predicted label after removing evidence from the source document. \textbf{Sufficiency} reflects the probability change of using only evidence to predict the label, in other words, if the evidence is really influential, the probability of the label will not change significantly. (2) The retrieved evidence needs to have \textbf{Plausibility} to convince the verifier's prediction \cite{deyoung2020eraser}. We adopt gold evidence to train the retrieved evidence. (3) The \textbf{Accuracy} of the task needs to be improved by revisiting the evidence retrieved.

\section{Model Architecture}

 
As shown in Figure \ref{fig:overview}, {\modelname} first leverage the labeled data to fine tune the claim verifier with $\mathcal{L}_{acc}$.{\modelname} utilizes gold evidence to provide plausibility of the retrieved evidence ($\mathcal{L}_{plau}$) and gold labels to provide faithfulness of evidence ($\mathcal{L}_{full}$ and $\mathcal{L}_{suff}$).

 
\subsection{Sentence Encoder}
We adopt the BERT encoder \cite{devlin2019bert} to obtain the semantic embeddings of each sentence within the claim and source document. For a given claim $C$ and its corresponding source document $D$, we get their sentence embeddings by adding a special token [CLS] at the beginning of each sentence and utilizing the [CLS] position embeddings. This produces an embedding matrix $S_{emb} \in \mathbb{R}^{l\times d}$ for the claim and document, where $l$ is number of total sentences and $d=768$.

\subsection{Claim Verifier}\label{sec:verifier}

Our claim verifier takes $S_{emb}$ as input and classifies the claim into three categories: refuted (Ref), supported (Sup) and not enough information (NEI). During training, the verifier performs classification based on the claim and the document. 

We use a neural network-based classifier $\mathcal{F}_{ver}$ to achieve this. It takes $S_{emb}$ as input and outputs a probability prediction vector $\mathcal{F}_{ver}(S_{emb}) = (p_{Sup}, p_{Ref}, p_{NEI})^\top$, where $p_{Sup}$, $p_{Ref}$ and $p_{NEI}$ represent the probability of claim Sup, Ref, or NEI, respectively. We denote the verification result as random variable $v$.

\subsubsection{Accuracy}
We adopt the criterion of accuracy to train the claim verifier to perform claim verification. To evaluate its performance, we use cross entropy loss $\mathcal{L}_{CE}(\mathcal{F}_{ver}(S_{emb}), y^*)$, which calculates the difference between the verifier's probability prediction $\mathcal{F}_{ver}(S_{emb})$ and the ground truth label $y^*\in\{0, 1, 2\}$ which indicates the Ref, Sup, and NEI, respectively. Consequently, we define the accuracy loss function as:
\begin{align}
    \mathcal{L}_{acc} = \mathcal{L}_{CE}(\mathcal{F}_{ver}(S_{emb}), y^*),
\end{align}
which is used to train the claim verifier and the sentence encoder. 

\subsection{Evidence Retriever}
After the claim verifier is trained, the evidence retriever will be trained to improve the faithfulness of the retrieved evidence using the trained verifier and ensure plausibility using the gold evidence in the dataset. The optimized evidence further enhances the performance of verification. To achieve this, we use a neural network-based classifier $\mathcal{F}_{ret}$ and the output of the sentence encoder to obtain semantic information. Notationally, $\mathcal{F}_{ret}$ takes $S_{emb}$ as input from the sentence encoder and outputs a vector $\mathcal{F}_{ret}(S_{emb}) \in [0,1]^{l}$, which quantifies the probability that each of the $l$ sentences in the document is important to claim verification. We denote $1, 0$ to indicate sentences are selected or not, respectively. We denote the sentence embedding obtained after passing the selected evidence to the sentence encoder as $E_{emb}$. 

To ensure faithfulness, we use the criteria of fullness and sufficiency. For more plausible evidence, we employ the criterion of plausibility, which incentivizes the retriever to have a evidence selection that makes sense to humans. We denote the loss function for fullness, sufficiency, plausibility as $\mathcal{L}_{full}$, $\mathcal{L}_{suff}$, and $\mathcal{L}_{plau}$ respectively. Consequently, we can use $\mathcal{L}$ to jointly represent the three loss functions as the target function for the evidence retriever:
\begin{align}
\mathcal{L} = \alpha_{full}\mathcal{L}_{full} + \alpha_{suff}\mathcal{L}_{suff} + \alpha_{plau}\mathcal{L}_{plau}.
\end{align}


\subsubsection{Plausibility} We introduce the plausibility criterion to measure and enhance the degree to which evidence is plausible to humans. To select the sentences that are most important to the claim verifier, we use a Top $k$ algorithm that selects the sentences with the highest probability scores. Specifically, we select the Top $k\%$ sentences in the document based on their probability scores. The selected evidence is denoted as $E$.

We adopt the claim with corresponding gold evidence and measure the difference between the predicted evidence and the gold evidence with binary cross entropy loss. We denote $\boldsymbol{g_i} \in \{0, 1\}^{|S|}$ as the gold evidence, where 0 or 1 represents whether a sentence is selected or not. The plausibility loss function could be defined as:
\begin{align}
    \mathcal{L}_{plau} = \mathcal{L}_{BCE}(\mathcal{F}_{ret}(S_{emb}), \boldsymbol{g_i}),
\end{align} 
which could encourage the retriever to select evidence sentences that are more plausible during training.

\subsubsection{Faithfulness-Fullness}
If removing certain sentences from the document would lead to incorrect verification result, we can assume that these sentences contain critical evidence that plays a crucial role in the verification outcome. To choose the most crucial evidence, we should identify the sentences that, if removed, would significantly reduce the claim verifier's performance.

We use cross entropy loss $\mathcal{L}_{CE}(\mathcal{F}_{ver}(S_{emb}), y^*)$ to measure the verification performance, where the label $y^*$ indicates one of three categories. To assess the impact of removing evidence sentences, we can compare the performance of ${S_{emb}\backslash E_{emb}}$ to the original input. Specifically, we can measure the influence of removing evidence sentences with the following formula:
\begin{align}
    \mathcal{L}_{full} = \mathcal{L}_{CE}(\mathcal{F}_{ver} (S_{emb}), y^*) - \mathcal{L}_{CE}(\mathcal{F}_{ver}({S_{emb}\backslash E_{emb}}, y^*).
\end{align}
The loss function $\mathcal{L}_{full}$ can encourage the retriever to select all sentences important to claim verification. 

Ideally, the evidence retriever selects the key evidence sentences that play an decisive part in the verification process so that $\mathcal{L}_{full} < 0$. To address this issue, we can first set $\mathcal{L'}_{full}$ to 0 when corresponding $\mathcal{L}_{full} < - B_f$, where $B_f > 0$ is a hyperparameter. To transform the range of the original loss values so that it is always 0 or more, we can denote $\mathcal{L'}_{full}$ = $\mathcal{L}_{full} + B_f$ when $\mathcal{L}_{full} > -B_f$ so that the reformulated loss value $\mathcal{L'}_{full} \geq 0$. Formally, we can define $\mathcal{L'}_{full}$ as follows:
\begin{align}
    \mathcal{L'}_{full} = \max(0, \mathcal{L}_{full} + B_f),
\end{align}
which could regulate the value of $\mathcal{L}_{full}$ into the range of $[0, +\infty)$.

\subsubsection{Faithfulness-Sufficiency}
To ensure that the selected evidence improves verification performance beyond what the original source document provides, we use the sufficiency criterion. This criterion incentivizes the retriever to select evidence that results in the greatest improvement in claim verification performance compared with using the original document alone.

More specifically, we adopt $\mathcal{L}_{CE}(\mathcal{F}_{ver} (E_{emb}), y^*)$ which represents the performance of using the evidence to replace the document, while $\mathcal{L}_{CE}(\mathcal{F}_{ver}(S_{emb}), y^*)$ stands for the original performance using the claim and the document as input to the claim verifier. Thus, we define the sufficiency loss function:
\begin{align}
    \mathcal{L}_{suff} = \mathcal{L}_{CE}(\mathcal{F}_{ver} (E_{emb}), y^*) - \mathcal{L}_{CE}(\mathcal{F}_{ver}(S_{emb}), y^*),
\end{align}
which encourages the retriever to select all important sentences that are used in the claim verification process.
The loss function $\mathcal{L}_{suff}$ also have the potential to be negative when the retriever is well-trained. To avoid a negative loss function, we can employ similar measurements by setting a hyperparameter $B_s > 0$, which is large enough and transforming the range of value into $[0, +\infty)$. Therefore, we can define the sufficiency loss function as:
\begin{align}
    \mathcal{L'}_{suff} = \max(0, \mathcal{L}_{suff} + B_s)
\end{align}

The optimized retriever will retrieve better evidence, which improves the results of the verifier in Section \ref{sec:verifier} by revisiting it.

\begin{table*}
\centering
\caption{Micro and Macro F1 Results of {\modelname} and baseline models across Test and Dev sets on CHEF.}
\vspace{-4mm}
\scalebox{0.78}{
\begin{tabular}{ccc|cccc|cccc}
\thickhline
\multicolumn{3}{c|}{\multirow{3}{*}{System / Evidence}} &  \multicolumn{4}{c|}{Test Set} &   \multicolumn{4}{c}{Dev Set}
\\ \cmidrule(lr){4-7} \cmidrule(lr){8-11}
& & &\multicolumn{2}{c}{BERT-Based Model} & \multicolumn{2}{c|}{RoBERTa-Based Model} & \multicolumn{2}{c}{BERT-Based Model} & \multicolumn{2}{c}{RoBERTa-Based Model} \\ \cmidrule(lr){4-5} \cmidrule(lr){6-7} \cmidrule(lr){8-9} \cmidrule(lr){10-11} 
& & & Micro F1 & Macro F1 & Micro F1 & Macro F1 & Micro F1 & Macro F1 & Micro F1 & Macro F1  \\
\midrule
\multirow{5}{*}{Pipeline} & \multicolumn{2}{|l|}{No Evidence} & 54.46\small±2.89 & 52.49\small±2.44 & 55.34\small±2.68 & 53.22\small±2.59 & 54.76\small±2.35 & 52.97\small±2.12 & 55.73\small±2.06 & 53.61\small±2.17\\
& \multicolumn{2}{|l|}{Google Snippets \cite{gupta2021x}} &\cellcolor{blue!15}62.07\small±2.55 &\cellcolor{blue!15}60.61\small±2.96  & \cellcolor{blue!15}62.53\small±2.13 & \cellcolor{blue!15}61.55\small±2.69  & \cellcolor{blue!15}62.31\small±1.97 & \cellcolor{blue!15}60.87\small±2.07 & \cellcolor{blue!15}62.96\small±2.17 & \cellcolor{blue!15}61.93\small±2.42 \\
& \multicolumn{2}{|l|}{Surface Ranker \cite{aly1feverous}}  & \cellcolor{green!15}63.17\small±1.67 & \cellcolor{green!15}61.47\small±2.02  & \cellcolor{green!15}64.21\small±1.94  & \cellcolor{green!15}62.05\small±2.17  & \cellcolor{green!15}63.53\small±1.78 & \cellcolor{green!15}61.78\small±1.95 & \cellcolor{green!15}64.66\small±1.86 & \cellcolor{green!15}62.49\small±2.08\\
& \multicolumn{2}{|l|}{Semantic Ranker \cite{Liu2020KernelGA}} & \cellcolor{green!15}63.47\small±1.71 & \cellcolor{green!15}61.94\small±1.66 & \cellcolor{green!15}64.35\small±1.76 & \cellcolor{green!15}62.24\small±1.52  &\cellcolor{green!15}63.73\small±1.68 & \cellcolor{green!15}62.42\small±1.49 &\cellcolor{green!15}64.71\small±1.45 & \cellcolor{green!15}62.59\small±1.38 \\
& \multicolumn{2}{|l|}{Hybrid Ranker \cite{Shaar2020ThatIA}} & \cellcolor{green!15}63.29\small±1.65  & \cellcolor{green!15}61.80\small±2.31  & \cellcolor{green!15}63.98\small±1.53 & \cellcolor{green!15}61.78\small±1.48 & \cellcolor{green!15}63.12\small±1.72 & \cellcolor{green!15}61.53\small±1.59& \cellcolor{green!15}64.32\small±1.83 & \cellcolor{green!15}62.11\small±1.43\\ 
\midrule
\multicolumn{1}{c}{\multirow{6}{*}{Joint}} & \multicolumn{1}{|l}{\multirow{2}{*}{Reinforce \cite{lei2016rationalizing}}} &
\multicolumn{1}{|l|}{Google Snippets} &\cellcolor{blue!15} 63.76\small±1.52 & \cellcolor{blue!15}61.74\small±1.88 &\cellcolor{blue!15}64.46\small±1.82 &\cellcolor{blue!15}62.42\small±1.67   &\cellcolor{blue!15}63.54\small±1.38 &\cellcolor{blue!15}61.48\small±1.63
&\cellcolor{blue!15}64.81\small±1.69 &\cellcolor{blue!15}62.80\small±1.72\\   
&\multicolumn{1}{|l}{} & \multicolumn{1}{|l|}{Source Documents} &\cellcolor{green!15}64.37\small±1.65 & \cellcolor{green!15}62.46\small±1.72 & \cellcolor{green!15}65.04\small±1.59 & \cellcolor{green!15}63.05\small±1.47   & \cellcolor{green!15}64.68\small±1.62 & \cellcolor{green!15}62.63\small±1.49
& \cellcolor{green!15}65.48\small±1.68 & \cellcolor{green!15}63.41\small±1.39 \\
\cmidrule(lr){2-3} 
 & \multicolumn{1}{|l}{\multirow{2}{*}{Multi-task \cite{Yin2018}}} &
\multicolumn{1}{|l|}{Google Snippets} &\cellcolor{blue!15} 62.78\small±1.41 & \cellcolor{blue!15}61.98\small±2.59 &\cellcolor{blue!15}64.19\small±1.98 &\cellcolor{blue!15}62.62\small±1.76   &\cellcolor{blue!15}62.94\small±1.86 &\cellcolor{blue!15}62.37\small±1.65
&\cellcolor{blue!15}64.51\small±1.79 &\cellcolor{blue!15}63.05\small±1.76 \\   
& \multicolumn{1}{|l}{} & \multicolumn{1}{|l|}{Source Documents} &\cellcolor{green!15} 65.02\small±1.46 & \cellcolor{green!15}63.12\small±1.78 & \cellcolor{green!15}65.87\small±1.68 & \cellcolor{green!15}63.79\small±1.84   & \cellcolor{green!15}65.41\small±1.80 & \cellcolor{green!15}63.38\small±1.62
 & \cellcolor{green!15}66.19\small±1.63 & \cellcolor{green!15}64.12\small±1.55\\
\cmidrule(lr){2-3} 
 &\multicolumn{1}{|l}{\multirow{2}{*}{Latent \cite{hu2022chef}}} &
\multicolumn{1}{|l|}{Google Snippets} &\cellcolor{blue!15} 64.45\small±1.68 & \cellcolor{blue!15}62.52\small±2.23 &\cellcolor{blue!15}65.11\small±1.86 &\cellcolor{blue!15}63.14\small±1.82   &\cellcolor{blue!15}64.71\small±1.69 &\cellcolor{blue!15}62.80\small±1.48
&\cellcolor{blue!15}65.08\small±1.62 &\cellcolor{blue!15}63.50\small±1.77\\ 
& \multicolumn{1}{|l}{} & \multicolumn{1}{|l|}{Source Documents} &\cellcolor{green!15}66.77\small±1.43 & \cellcolor{green!15}64.65\small±1.74 & \cellcolor{green!15}66.95\small±1.68 & \cellcolor{green!15}65.13\small±1.57   & \cellcolor{green!15}66.96\small±1.45 & \cellcolor{green!15}64.92\small±1.50  & \cellcolor{green!15}67.33\small±1.26 & \cellcolor{green!15}65.57\small±1.39  \\
\midrule
\multicolumn{1}{c}{\multirow{3}{*}{Pipeline}} & \multicolumn{1}{|l}{\textbf{\modelname}} & \multicolumn{1}{|l|}{Source Documents} &\cellcolor{green!15} \textbf{70.87\small±1.05} & \cellcolor{green!15}\textbf{68.78\small±1.21} & \cellcolor{green!15}\textbf{71.24\small±1.11} & \cellcolor{green!15}\textbf{69.52\small±0.96}   & \cellcolor{green!15}\textbf{71.31\small±1.08} & \cellcolor{green!15}\textbf{69.25\small±1.18}
 & \cellcolor{green!15}\textbf{71.79\small±1.26} & \cellcolor{green!15}\textbf{69.98\small±1.09} \\
& \multicolumn{1}{|l}{\textit{w/o $\mathcal{L}_{plau}$}} & \multicolumn{1}{|l|}{Source Documents} &\cellcolor{green!15}67.67\small±1.32 & \cellcolor{green!15}65.84\small±1.46 & \cellcolor{green!15}68.03\small±1.35 & \cellcolor{green!15}66.11\small±1.48   & \cellcolor{green!15}67.96\small±1.57 & \cellcolor{green!15}66.04\small±1.51
 & \cellcolor{green!15}68.14\small±1.42 & \cellcolor{green!15}66.31\small±1.56 \\
& \multicolumn{1}{|l}{\textit{w/o $\mathcal{L}_{full}$\&$\mathcal{L}_{suff}$}} & \multicolumn{1}{|l|}{Source Documents} &\cellcolor{green!15}68.24\small±1.42 & \cellcolor{green!15}66.15\small±1.39 & \cellcolor{green!15}68.58\small±1.50 & \cellcolor{green!15}66.39\small±1.44   & \cellcolor{green!15}68.53\small±1.32 & \cellcolor{green!15}66.31\small±1.53
 & \cellcolor{green!15}68.70\small±1.44 & \cellcolor{green!15}66.59\small±1.37 \\
\midrule
\midrule
\multirow{1}{*}{Pipeline} &
\multicolumn{2}{|l|}{Gold Evidence} & \textbf{78.99\small±0.82} & \textbf{77.62\small±1.02}  & \textbf{79.14\small±0.93} & \textbf{78.59\small±1.02} & \textbf{79.26\small±0.94} & \textbf{78.04\small±1.10}& \textbf{79.98\small±0.89} & \textbf{78.81\small±1.01}   \\

\thickhline
\end{tabular}}
\label{tab:verification}
\vspace{-4mm}
\end{table*}

\vspace{-2mm}
\section{Experiments and Analyses}
\label{sec:experiments}

\subsection{Setup and Baselines}
\paragraph{Setup:}
Note that only CHEF \cite{hu2022chef} has marked the gold evidence for real-world claims. Although FEVER, \cite{thorne2018fever}, FEVER 2.0 \cite{Thorne19FEVER2}, and FEVEROUS \cite{aly1feverous} annotate evidence retrieved from Wikipedia, they do not serve claims from the real-world. Therefore, we only use CHEF. To measure the effect of {\modelname}, we adjust the parameters on the train set, and report the results on dev and test sets of CHEF. The train/dev/test sets of CHEF have 8,002/999/999 samples respectively.
CHEF also provides the google snippets as the evidence, which is the summary of the content of source documents provided by Google \cite{gupta2021x}.
Following prior efforts \cite{hu2022chef,gupta2021x,Liu2020KernelGA}, we adopt Micro F1 and Macro F1 as the evaluation metric. For base encoder, we adopt BERT-Base-Chinese \cite{devlin2019bert} and RoBERTa-Base-Chinese \cite{liu2019roberta}.
We set $k$ as 5\% of all sentences in the source documents. We use BertAdam~\citep{kingma2014adam} with 4e-5 learning rate, warmup with 0.07 to optimize the cross entropy loss and set the batch size as 16. For simplicity, we set $\alpha_{full}$, $\alpha_{suff}$, and $\alpha_{plau}$ to 1 respectively.

\vspace{-2mm}
\paragraph{Baselines:}
Following previous works \cite{hu2022chef,gupta2021x}, we adopt two types of baselines: Pipeline and Joint systems. Pipeline systems first retrieve evidence from the documents according to the claim, and use the retrieved evidence to verify the claim. The evidence retriever and claim verification are two independent steps. We adopt (1) Google Snippets \cite{gupta2021x}. (2) Surface Ranker \cite{aly1feverous}. (3) Semantic Ranker \cite{Liu2020KernelGA}. (4) Hybrid Ranker \cite{Shaar2020ThatIA}. Joint systems treat evidence extraction as a latent variable, and jointly optimize the evidence extraction process by claim verification loss. We adopt (5) Reinforcement-based Method \cite{lei2016rationalizing}. (6) Multi-task based Method \cite{Yin2018}. (7) Latent based Method \cite{hu2022chef}. In addition, we give (8) No evidence and (9) Gold evidence, to show lower and upper bounds for results.

\vspace{-2mm}
\subsection{Results and Analysis}

\noindent \textbf{Overall Performance.}
Table \ref{tab:verification} shows the mean and standard deviation results with 5 runs of training and testing on dev and test sets of CHEF. We observe that the use of real-world evidence can improve the effect of claim verification, and source documents can bring more improvement than google snippets, which is related to the fact that source documents contains more information. Correspondingly, these source documents also contain more noise content, but {\modelname} still consistently outperforms the baselines.  More specifically, compared with the previous SOTA model: Latent \cite{hu2022chef}, {\modelname} on average achieves 4.30\% higher Micro F1 and 4.32\% higher Macro F1 across dev and test sets. We attribute the consistent improvement of {\modelname} to the faithful and plausible evidence which {\modelname} retrieved from source documents.
{\modelname} is more robust than all baselines when considering standard deviations, since the evidence retriever is supervised by gold evidence through plausibility, providing higher quality evidence.

\begin{table}[t]
\centering
\caption{Quality of Retrieved Evidence Analysis.}
\vspace{-4mm}
\scalebox{0.77}{
\begin{tabular}{l|cccc|cccc}
\thickhline
\multicolumn{1}{c|}{\multirow{3}{*}{Methods}} & \multicolumn{4}{c|}{Test Set} & \multicolumn{4}{c}{Dev Set}
\\ \cmidrule(lr){2-5} \cmidrule(lr){6-9} 
& \multicolumn{2}{c}{BERT-Base} & \multicolumn{2}{c|}{RoBERTa-Base} &  \multicolumn{2}{c}{BERT-Base} & \multicolumn{2}{c}{RoBERTa-Base}
\\ \cmidrule(lr){2-3} \cmidrule(lr){4-5} \cmidrule(lr){6-7} \cmidrule(lr){8-9}
&BLEU & F1  & BLEU   & F1 & BLEU   & F1 & BLEU   & F1   \\
\midrule 
\multicolumn{1}{l|}{Surface} & 0.43 & 85.3  &  0.46  & 86.6 &  0.42 &  84.6 &  0.44 &  85.5\\
\multicolumn{1}{l|}{Semantic} & 0.53 & 88.1 &  0.55  & 89.5  &  0.52 & 88.4 & 0.56 &  89.4\\
\multicolumn{1}{l|}{Hybrid} & 0.48  & 87.7 & 0.50  &  88.9 & 0.46 &  87.5 & 0.48 & 88.6\\
\multicolumn{1}{l|}{Reinforce} & 0.63 & 89.6 & 0.66  & 90.4 & 0.62 & 89.3 & 0.64  & 90.3\\
\multicolumn{1}{l|}{Multi-task} &  0.66 & 90.4 &  0.67 & 91.5 & 0.64 &  90.3 &  0.65& 90.8\\
\multicolumn{1}{l|}{Latent} &  0.68 & 90.8 &  0.69 & 91.4 & 0.67 &  90.5 &  0.69 & 91.2\\
\multicolumn{1}{l|}{\textbf{{\modelname}}} &\textbf{0.84} &\textbf{95.3} & \textbf{0.86} & \textbf{95.4} &\textbf{0.85} & \textbf{95.1} &\textbf{0.87} & \textbf{95.7}\\
\thickhline
\end{tabular}}
\label{tab:human}
\vspace{-5mm}
\end{table}

\noindent \textbf{Ablation Study.}
We conduct an ablation study to show the effectiveness of different losses of {\modelname} on the dev and test sets. {\modelname} \textit{w/o $\mathcal{L}_{plau}$} means that the plausible loss function is removed, which makes the evidence retriever no longer use the gold evidence to train the selected evidence. {\modelname} \textit{w/o $\mathcal{L}_{full}$\&$\mathcal{L}_{suff}$} removes the faithful loss function from the claim verifier, which will cause the evidence obtained by the evidence retriever to no longer depend on the claim verification result. 
A general conclusion from ablation rows in Table \ref{tab:verification} is that all losses contribute positively to the improved performance. More specifically, without $\mathcal{L}_{plau}$, the selected evidence will become unconvincing, resulting in a 3.33\% F1 performance decrease. Removing the $\mathcal{L}_{full}$\&$\mathcal{L}_{suff}$ will select task-agnostic evidence, resulting in a 2.90\% F1 performance loss.

\noindent \textbf{Quality of Retrieved Evidence Analysis.}
We assess the retrieved evidence quality by comparing it to gold evidence in dev and test sets. We use the BLEU \cite{papineni2002bleu} to gauge the similarity between retrieved and gold evidence, with higher BLEU indicating better quality. Additionally, 5 Ph.D. students annotate verification labels for 100 claims based on retrieved evidence, while 2 Ph.D. students validate the data. This helps us evaluate the \textbf{interpretability} of retrieved evidence. Table \ref{tab:human} displays the BLEU and Micro F1 scores. {\modelname} shows a notable 17\% BLEU improvement over the SOTA baseline, proving that incorporating plausible loss for evidence retriever training helps {\modelname} obtain higher-quality evidence, resulting in a 5.87\% increase in human-labeled F1 verification accuracy.

\begin{figure}
    \centering  \includegraphics[scale=0.38]{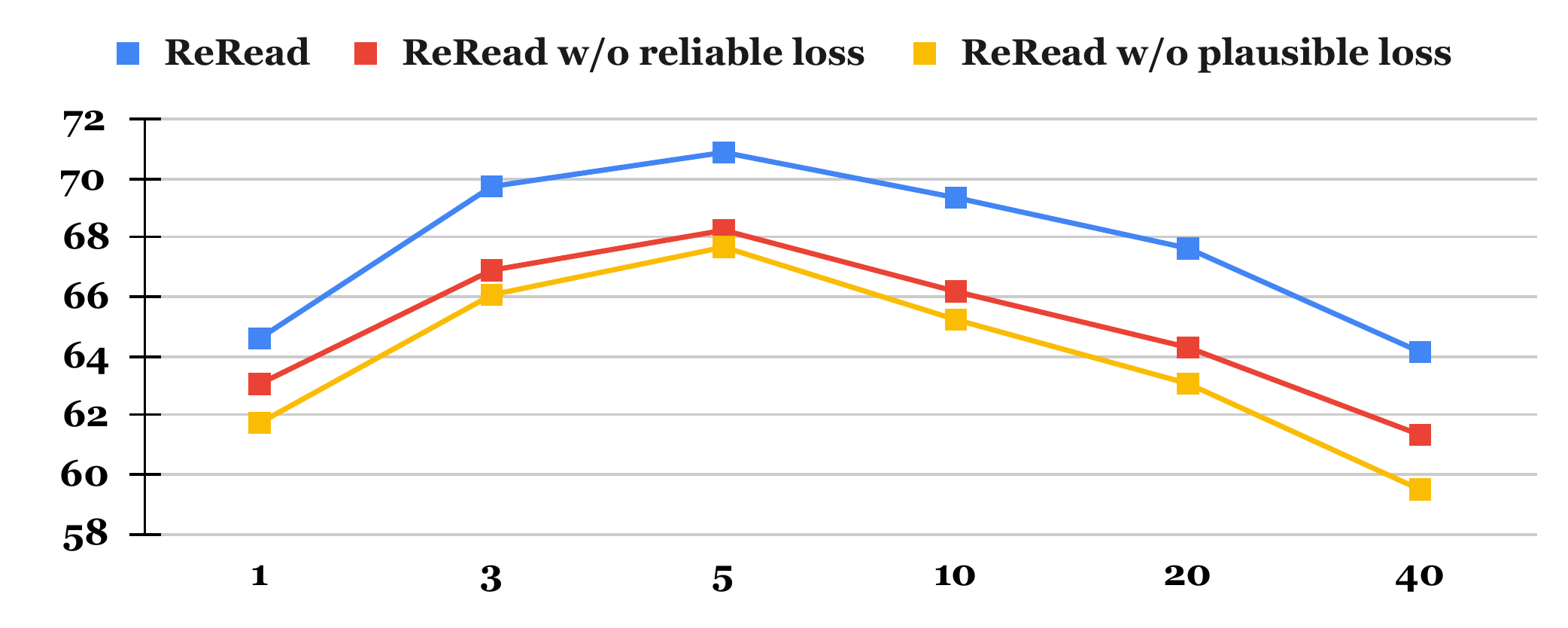}
    \vspace{-4mm}
    \caption{Micro F1 results with different $k$ on test set.}
    \label{fig:factorK}
    \vspace{-5mm}
\end{figure}

\noindent \textbf{Effect of the Selection Ratio $k$.}
As shown in Figure~\ref{fig:factorK}, we report Micro F1 scores of BERT-Base encoder against different $k$ on the test set. A low $k$ value may have a detrimental effect on the information sufficiency of the retrieved evidence, thus affecting the verification results. 
The F1 score of {\modelname} does not increase monotonically, as irrelevant evidence are included. The model achieves the best performance when $k=5$, which means 5\% sentences are selected as evidence is the most appropriate. If we remove the faithful and plausible loss, the F1 performance of {\modelname} will drop 3.24\% F1 on average due to missing guidance from the gold label and evidence.

\vspace{-2mm}
\section{Conclusion}
\label{conclusion}

In this paper, we propose a novel fact verification framework {\modelname}, which adopt the plausibility, fullness, and sufficiency criteria to retrieve appropriate evidence from real-world documents. The retrieved evidence could reflect the factuality of the claim and convince to human. With the training of the evidence retriever, it can further provide the claim verifier with better evidence to revisit and improve the accuracy of the verification task. Experiments on real-world dataset shows the effectiveness of {\modelname}. In the future, we can extend the research on faithful interpretation to the construction of knowledge graphs \cite{hu2020selfore,hu2021gradient,hu2020semi,liu2022hierarchical,zhang2022domain}, the extraction and answering of structured knowledge \cite{liu2023comprehensive,liu2022semantic}.

\begin{acks}
Lijie Wen is the corresponding author. Xuming Hu and Zhaochen Hong have made equal contributions to this work. Our code is available at: \url{https://github.com/THU-BPM/ReRead}. The work was supported by the National Key Research and Development Program of China (No. 2019YFB1704003), the National Nature Science Foundation of China (No. 62021002), NSF under grants III1909323, Tsinghua BNRist and Beijing Key Laboratory of Industrial Bigdata System and Application.
\end{acks}
\balance
\bibliographystyle{ACM-Reference-Format}
\bibliography{sample-base}


\begin{thebibliography}{37}


\ifx \showCODEN    \undefined \def \showCODEN     #1{\unskip}     \fi
\ifx \showDOI      \undefined \def \showDOI       #1{#1}\fi
\ifx \showISBNx    \undefined \def \showISBNx     #1{\unskip}     \fi
\ifx \showISBNxiii \undefined \def \showISBNxiii  #1{\unskip}     \fi
\ifx \showISSN     \undefined \def \showISSN      #1{\unskip}     \fi
\ifx \showLCCN     \undefined \def \showLCCN      #1{\unskip}     \fi
\ifx \shownote     \undefined \def \shownote      #1{#1}          \fi
\ifx \showarticletitle \undefined \def \showarticletitle #1{#1}   \fi
\ifx \showURL      \undefined \def \showURL       {\relax}        \fi
\providecommand\bibfield[2]{#2}
\providecommand\bibinfo[2]{#2}
\providecommand\natexlab[1]{#1}
\providecommand\showeprint[2][]{arXiv:#2}

\bibitem[Adair et~al\mbox{.}(2017)]%
        {Adair2017ProgressT}
\bibfield{author}{\bibinfo{person}{Bill Adair}, \bibinfo{person}{Chengkai Li},
  \bibinfo{person}{Jun Yang}, {and} \bibinfo{person}{Cong Yu}.}
  \bibinfo{year}{2017}\natexlab{}.
\newblock \showarticletitle{Progress toward “the holy grail”: The continued
  quest to automate fact-checking}. In \bibinfo{booktitle}{\emph{Proceedings of
  the 2017 Computation+Journalism Symposium}}.
\newblock


\bibitem[Aly et~al\mbox{.}(2021)]%
        {aly1feverous}
\bibfield{author}{\bibinfo{person}{Rami Aly}, \bibinfo{person}{Zhijiang Guo},
  \bibinfo{person}{Michael~Sejr Schlichtkrull}, \bibinfo{person}{James Thorne},
  \bibinfo{person}{Andreas Vlachos}, \bibinfo{person}{Christos
  Christodoulopoulos}, \bibinfo{person}{Oana Cocarascu}, {and}
  \bibinfo{person}{Arpit Mittal}.} \bibinfo{year}{2021}\natexlab{}.
\newblock \showarticletitle{FEVEROUS: Fact Extraction and VERification Over
  Unstructured and Structured information}. In
  \bibinfo{booktitle}{\emph{Thirty-fifth Conference on Neural Information
  Processing Systems Datasets and Benchmarks Track (Round 1)}}.
\newblock


\bibitem[Botnevik et~al\mbox{.}(2020)]%
        {botnevik2020brenda}
\bibfield{author}{\bibinfo{person}{Bjarte Botnevik}, \bibinfo{person}{Eirik
  Sakariassen}, {and} \bibinfo{person}{Vinay Setty}.}
  \bibinfo{year}{2020}\natexlab{}.
\newblock \showarticletitle{Brenda: Browser extension for fake news detection}.
  In \bibinfo{booktitle}{\emph{Proc. of SIGIR}}. \bibinfo{pages}{2117--2120}.
\newblock


\bibitem[Chen et~al\mbox{.}(2022)]%
        {chen2022gere}
\bibfield{author}{\bibinfo{person}{Jiangui Chen}, \bibinfo{person}{Ruqing
  Zhang}, \bibinfo{person}{Jiafeng Guo}, \bibinfo{person}{Yixing Fan}, {and}
  \bibinfo{person}{Xueqi Cheng}.} \bibinfo{year}{2022}\natexlab{}.
\newblock \showarticletitle{GERE: Generative evidence retrieval for fact
  verification}. In \bibinfo{booktitle}{\emph{Proc. of SIGIR}}.
  \bibinfo{pages}{2184--2189}.
\newblock


\bibitem[Devlin et~al\mbox{.}(2019)]%
        {devlin2019bert}
\bibfield{author}{\bibinfo{person}{Jacob Devlin}, \bibinfo{person}{Ming-Wei
  Chang}, \bibinfo{person}{Kenton Lee}, {and} \bibinfo{person}{Kristina
  Toutanova}.} \bibinfo{year}{2019}\natexlab{}.
\newblock \showarticletitle{BERT: Pre-training of Deep Bidirectional
  Transformers for Language Understanding}. In \bibinfo{booktitle}{\emph{Proc.
  of NAACL-HLT}}. \bibinfo{pages}{4171--4186}.
\newblock


\bibitem[DeYoung et~al\mbox{.}(2020)]%
        {deyoung2020eraser}
\bibfield{author}{\bibinfo{person}{Jay DeYoung}, \bibinfo{person}{Sarthak
  Jain}, \bibinfo{person}{Nazneen~Fatema Rajani}, \bibinfo{person}{Eric
  Lehman}, \bibinfo{person}{Caiming Xiong}, \bibinfo{person}{Richard Socher},
  {and} \bibinfo{person}{Byron~C Wallace}.} \bibinfo{year}{2020}\natexlab{}.
\newblock \showarticletitle{ERASER: A Benchmark to Evaluate Rationalized NLP
  Models}. In \bibinfo{booktitle}{\emph{Proc. ACL}}.
  \bibinfo{pages}{4443--4458}.
\newblock


\bibitem[Graves(2018)]%
        {2018graves}
\bibfield{author}{\bibinfo{person}{Lucas Graves}.}
  \bibinfo{year}{2018}\natexlab{}.
\newblock \showarticletitle{Understanding the Promise and Limits of Automated
  Fact-checking}.
\newblock \bibinfo{journal}{\emph{Reuters Institute for the Study of
  Journalism}} (\bibinfo{year}{2018}).
\newblock


\bibitem[Guo et~al\mbox{.}(2022)]%
        {guo2022survey}
\bibfield{author}{\bibinfo{person}{Zhijiang Guo}, \bibinfo{person}{Michael
  Schlichtkrull}, {and} \bibinfo{person}{Andreas Vlachos}.}
  \bibinfo{year}{2022}\natexlab{}.
\newblock \showarticletitle{A survey on automated fact-checking}.
\newblock \bibinfo{journal}{\emph{Transactions of the Association for
  Computational Linguistics}}  \bibinfo{volume}{10} (\bibinfo{year}{2022}),
  \bibinfo{pages}{178--206}.
\newblock


\bibitem[Gupta and Srikumar(2021)]%
        {gupta2021x}
\bibfield{author}{\bibinfo{person}{Ashim Gupta} {and} \bibinfo{person}{Vivek
  Srikumar}.} \bibinfo{year}{2021}\natexlab{}.
\newblock \showarticletitle{X-Fact: A New Benchmark Dataset for Multilingual
  Fact Checking}. In \bibinfo{booktitle}{\emph{Proc. of ACL}}.
  \bibinfo{pages}{675--682}.
\newblock


\bibitem[Hu et~al\mbox{.}(2022)]%
        {hu2022chef}
\bibfield{author}{\bibinfo{person}{Xuming Hu}, \bibinfo{person}{Zhijiang Guo},
  \bibinfo{person}{GuanYu Wu}, \bibinfo{person}{Aiwei Liu},
  \bibinfo{person}{Lijie Wen}, {and} \bibinfo{person}{S~Yu Philip}.}
  \bibinfo{year}{2022}\natexlab{}.
\newblock \showarticletitle{CHEF: A Pilot Chinese Dataset for Evidence-Based
  Fact-Checking}. In \bibinfo{booktitle}{\emph{Proc. of NAACL-HLT}}.
  \bibinfo{pages}{3362--3376}.
\newblock


\bibitem[Hu et~al\mbox{.}(2020)]%
        {hu2020selfore}
\bibfield{author}{\bibinfo{person}{Xuming Hu}, \bibinfo{person}{Lijie Wen},
  \bibinfo{person}{Yusong Xu}, \bibinfo{person}{Chenwei Zhang}, {and}
  \bibinfo{person}{Philip Yu}.} \bibinfo{year}{2020}\natexlab{}.
\newblock \showarticletitle{{S}elf{ORE}: Self-supervised Relational Feature
  Learning for Open Relation Extraction}. In \bibinfo{booktitle}{\emph{Proc. of
  EMNLP}}. \bibinfo{address}{Online}, \bibinfo{pages}{3673--3682}.
\newblock


\bibitem[Hu et~al\mbox{.}(2021a)]%
        {hu2020semi}
\bibfield{author}{\bibinfo{person}{Xuming Hu}, \bibinfo{person}{Chenwei Zhang},
  \bibinfo{person}{Fukun Ma}, \bibinfo{person}{Chenyao Liu},
  \bibinfo{person}{Lijie Wen}, {and} \bibinfo{person}{S~Yu Philip}.}
  \bibinfo{year}{2021}\natexlab{a}.
\newblock \showarticletitle{Semi-supervised Relation Extraction via Incremental
  Meta Self-Training}. In \bibinfo{booktitle}{\emph{Findings of EMNLP}}.
  \bibinfo{pages}{487--496}.
\newblock


\bibitem[Hu et~al\mbox{.}(2021b)]%
        {hu2021gradient}
\bibfield{author}{\bibinfo{person}{Xuming Hu}, \bibinfo{person}{Chenwei Zhang},
  \bibinfo{person}{Yawen Yang}, \bibinfo{person}{Xiaohe Li},
  \bibinfo{person}{Li Lin}, \bibinfo{person}{Lijie Wen}, {and}
  \bibinfo{person}{S~Yu Philip}.} \bibinfo{year}{2021}\natexlab{b}.
\newblock \showarticletitle{Gradient Imitation Reinforcement Learning for Low
  Resource Relation Extraction}. In \bibinfo{booktitle}{\emph{Proc. of EMNLP}}.
  \bibinfo{pages}{2737--2746}.
\newblock


\bibitem[Jacovi and Goldberg(2020)]%
        {jacovi2020towards}
\bibfield{author}{\bibinfo{person}{Alon Jacovi} {and} \bibinfo{person}{Yoav
  Goldberg}.} \bibinfo{year}{2020}\natexlab{}.
\newblock \showarticletitle{Towards Faithfully Interpretable {NLP} Systems: How
  Should We Define and Evaluate Faithfulness?}. In
  \bibinfo{booktitle}{\emph{Proceedings of the 58th Annual Meeting of the
  Association for Computational Linguistics}}. \bibinfo{publisher}{Association
  for Computational Linguistics}, \bibinfo{address}{Online},
  \bibinfo{pages}{4198--4205}.
\newblock
\urldef\tempurl%
\url{https://doi.org/10.18653/v1/2020.acl-main.386}
\showDOI{\tempurl}


\bibitem[Kingma and Ba(2015)]%
        {kingma2014adam}
\bibfield{author}{\bibinfo{person}{Diederik~P Kingma} {and}
  \bibinfo{person}{Jimmy Ba}.} \bibinfo{year}{2015}\natexlab{}.
\newblock \showarticletitle{Adam: A method for stochastic optimization}. In
  \bibinfo{booktitle}{\emph{Proc. of ICLR}}.
\newblock


\bibitem[Lei et~al\mbox{.}(2016)]%
        {lei2016rationalizing}
\bibfield{author}{\bibinfo{person}{Tao Lei}, \bibinfo{person}{Regina Barzilay},
  {and} \bibinfo{person}{Tommi Jaakkola}.} \bibinfo{year}{2016}\natexlab{}.
\newblock \showarticletitle{Rationalizing Neural Predictions}. In
  \bibinfo{booktitle}{\emph{Proc. of EMNLP}}. \bibinfo{pages}{107--117}.
\newblock


\bibitem[Liu et~al\mbox{.}(2022a)]%
        {liu2022semantic}
\bibfield{author}{\bibinfo{person}{Aiwei Liu}, \bibinfo{person}{Xuming Hu},
  \bibinfo{person}{Li Lin}, {and} \bibinfo{person}{Lijie Wen}.}
  \bibinfo{year}{2022}\natexlab{a}.
\newblock \showarticletitle{Semantic Enhanced Text-to-SQL Parsing via
  Iteratively Learning Schema Linking Graph}. In
  \bibinfo{booktitle}{\emph{Proc. of KDD}}. \bibinfo{pages}{1021--1030}.
\newblock


\bibitem[Liu et~al\mbox{.}(2023)]%
        {liu2023comprehensive}
\bibfield{author}{\bibinfo{person}{Aiwei Liu}, \bibinfo{person}{Xuming Hu},
  \bibinfo{person}{Lijie Wen}, {and} \bibinfo{person}{Philip~S Yu}.}
  \bibinfo{year}{2023}\natexlab{}.
\newblock \showarticletitle{A comprehensive evaluation of ChatGPT's zero-shot
  Text-to-SQL capability}.
\newblock \bibinfo{journal}{\emph{arXiv preprint arXiv:2303.13547}}
  (\bibinfo{year}{2023}).
\newblock


\bibitem[Liu et~al\mbox{.}(2022b)]%
        {liu2022hierarchical}
\bibfield{author}{\bibinfo{person}{Shuliang Liu}, \bibinfo{person}{Xuming Hu},
  \bibinfo{person}{Chenwei Zhang}, \bibinfo{person}{Shu'ang Li},
  \bibinfo{person}{Lijie Wen}, {and} \bibinfo{person}{Philip~S. Yu}.}
  \bibinfo{year}{2022}\natexlab{b}.
\newblock \showarticletitle{HiURE: Hierarchical Exemplar Contrastive Learning
  for Unsupervised Relation Extraction}. In \bibinfo{booktitle}{\emph{Proc. of
  NAACL-HLT}}. \bibinfo{pages}{5970--5980}.
\newblock


\bibitem[Liu et~al\mbox{.}(2019)]%
        {liu2019roberta}
\bibfield{author}{\bibinfo{person}{Yinhan Liu}, \bibinfo{person}{Myle Ott},
  \bibinfo{person}{Naman Goyal}, \bibinfo{person}{Jingfei Du},
  \bibinfo{person}{Mandar Joshi}, \bibinfo{person}{Danqi Chen},
  \bibinfo{person}{Omer Levy}, \bibinfo{person}{Mike Lewis},
  \bibinfo{person}{Luke Zettlemoyer}, {and} \bibinfo{person}{Veselin
  Stoyanov}.} \bibinfo{year}{2019}\natexlab{}.
\newblock \showarticletitle{Roberta: A robustly optimized bert pretraining
  approach}.
\newblock \bibinfo{journal}{\emph{arXiv preprint arXiv:1907.11692}}
  (\bibinfo{year}{2019}).
\newblock


\bibitem[Liu et~al\mbox{.}(2020)]%
        {Liu2020KernelGA}
\bibfield{author}{\bibinfo{person}{Zhenghao Liu}, \bibinfo{person}{Chenyan
  Xiong}, \bibinfo{person}{Maosong Sun}, {and} \bibinfo{person}{Zhiyuan Liu}.}
  \bibinfo{year}{2020}\natexlab{}.
\newblock \showarticletitle{Fine-grained Fact Verification with Kernel Graph
  Attention Network}. In \bibinfo{booktitle}{\emph{Proc. of ACL}}.
  \bibinfo{publisher}{Association for Computational Linguistics},
  \bibinfo{address}{Online}, \bibinfo{pages}{7342--7351}.
\newblock


\bibitem[Nielsen and McConville(2022)]%
        {nielsen2022mumin}
\bibfield{author}{\bibinfo{person}{Dan~S Nielsen} {and} \bibinfo{person}{Ryan
  McConville}.} \bibinfo{year}{2022}\natexlab{}.
\newblock \showarticletitle{Mumin: A large-scale multilingual multimodal
  fact-checked misinformation social network dataset}. In
  \bibinfo{booktitle}{\emph{Proc. of SIGIR}}. \bibinfo{pages}{3141--3153}.
\newblock


\bibitem[Papineni et~al\mbox{.}(2002)]%
        {papineni2002bleu}
\bibfield{author}{\bibinfo{person}{Kishore Papineni}, \bibinfo{person}{Salim
  Roukos}, \bibinfo{person}{Todd Ward}, {and} \bibinfo{person}{Wei-Jing Zhu}.}
  \bibinfo{year}{2002}\natexlab{}.
\newblock \showarticletitle{Bleu: a method for automatic evaluation of machine
  translation}. In \bibinfo{booktitle}{\emph{Proc. of ACL}}.
  \bibinfo{pages}{311--318}.
\newblock


\bibitem[Pradeep et~al\mbox{.}(2021)]%
        {pradeep2021vera}
\bibfield{author}{\bibinfo{person}{Ronak Pradeep}, \bibinfo{person}{Xueguang
  Ma}, \bibinfo{person}{Rodrigo Nogueira}, {and} \bibinfo{person}{Jimmy Lin}.}
  \bibinfo{year}{2021}\natexlab{}.
\newblock \showarticletitle{Vera: Prediction techniques for reducing harmful
  misinformation in consumer health search}. In \bibinfo{booktitle}{\emph{Proc.
  of SIGIR}}. \bibinfo{pages}{2066--2070}.
\newblock


\bibitem[Rashkin et~al\mbox{.}(2017)]%
        {Rashkin2017TruthOV}
\bibfield{author}{\bibinfo{person}{Hannah Rashkin}, \bibinfo{person}{Eunsol
  Choi}, \bibinfo{person}{Jin~Yea Jang}, \bibinfo{person}{Svitlana Volkova},
  {and} \bibinfo{person}{Yejin Choi}.} \bibinfo{year}{2017}\natexlab{}.
\newblock \showarticletitle{Truth of Varying Shades: Analyzing Language in Fake
  News and Political Fact-Checking}. In \bibinfo{booktitle}{\emph{Proc. of
  EMNLP}}. \bibinfo{publisher}{Association for Computational Linguistics},
  \bibinfo{address}{Copenhagen, Denmark}, \bibinfo{pages}{2931--2937}.
\newblock


\bibitem[Schuster et~al\mbox{.}(2020)]%
        {Schuster2020TheLO}
\bibfield{author}{\bibinfo{person}{Tal Schuster}, \bibinfo{person}{Roei
  Schuster}, \bibinfo{person}{Darsh~J. Shah}, {and} \bibinfo{person}{Regina
  Barzilay}.} \bibinfo{year}{2020}\natexlab{}.
\newblock \showarticletitle{The Limitations of Stylometry for Detecting
  Machine-Generated Fake News}.
\newblock \bibinfo{journal}{\emph{Computational Linguistics}}
  \bibinfo{volume}{46}, \bibinfo{number}{2} (\bibinfo{year}{2020}),
  \bibinfo{pages}{499--510}.
\newblock


\bibitem[Shaar et~al\mbox{.}(2020)]%
        {Shaar2020ThatIA}
\bibfield{author}{\bibinfo{person}{Shaden Shaar}, \bibinfo{person}{Nikolay
  Babulkov}, \bibinfo{person}{Giovanni Da~San~Martino}, {and}
  \bibinfo{person}{Preslav Nakov}.} \bibinfo{year}{2020}\natexlab{}.
\newblock \showarticletitle{That is a Known Lie: Detecting Previously
  Fact-Checked Claims}. In \bibinfo{booktitle}{\emph{Proc. of ACL}}.
  \bibinfo{publisher}{Association for Computational Linguistics},
  \bibinfo{address}{Online}, \bibinfo{pages}{3607--3618}.
\newblock


\bibitem[Thorne et~al\mbox{.}(2018a)]%
        {Thorne2018FEVERAL}
\bibfield{author}{\bibinfo{person}{James Thorne}, \bibinfo{person}{Andreas
  Vlachos}, \bibinfo{person}{Christos Christodoulopoulos}, {and}
  \bibinfo{person}{Arpit Mittal}.} \bibinfo{year}{2018}\natexlab{a}.
\newblock \showarticletitle{{FEVER}: a Large-scale Dataset for Fact Extraction
  and {VER}ification}. In \bibinfo{booktitle}{\emph{Proc. of NAACL-HLT}}.
  \bibinfo{publisher}{Association for Computational Linguistics},
  \bibinfo{address}{New Orleans, Louisiana}, \bibinfo{pages}{809--819}.
\newblock


\bibitem[Thorne et~al\mbox{.}(2018b)]%
        {thorne2018fever}
\bibfield{author}{\bibinfo{person}{James Thorne}, \bibinfo{person}{Andreas
  Vlachos}, \bibinfo{person}{Christos Christodoulopoulos}, {and}
  \bibinfo{person}{Arpit Mittal}.} \bibinfo{year}{2018}\natexlab{b}.
\newblock \showarticletitle{FEVER: a Large-scale Dataset for Fact Extraction
  and VERification}. In \bibinfo{booktitle}{\emph{Proc. NAACL-HLT}}.
  \bibinfo{pages}{809--819}.
\newblock


\bibitem[Thorne et~al\mbox{.}(2018c)]%
        {Thorne19FEVER2}
\bibfield{author}{\bibinfo{person}{James Thorne}, \bibinfo{person}{Andreas
  Vlachos}, \bibinfo{person}{Oana Cocarascu}, \bibinfo{person}{Christos
  Christodoulopoulos}, {and} \bibinfo{person}{Arpit Mittal}.}
  \bibinfo{year}{2018}\natexlab{c}.
\newblock \showarticletitle{The {FEVER2.0} Shared Task}. In
  \bibinfo{booktitle}{\emph{Proceedings of the Second Workshop on {Fact
  Extraction and VERification (FEVER)}}}.
\newblock


\bibitem[Turenne(2018)]%
        {turenne2018rumour}
\bibfield{author}{\bibinfo{person}{Nicolas Turenne}.}
  \bibinfo{year}{2018}\natexlab{}.
\newblock \showarticletitle{The rumour spectrum}.
\newblock \bibinfo{journal}{\emph{PloS one}} \bibinfo{volume}{13},
  \bibinfo{number}{1} (\bibinfo{year}{2018}), \bibinfo{pages}{e0189080}.
\newblock


\bibitem[Vo and Lee(2018)]%
        {vo2018rise}
\bibfield{author}{\bibinfo{person}{Nguyen Vo} {and} \bibinfo{person}{Kyumin
  Lee}.} \bibinfo{year}{2018}\natexlab{}.
\newblock \showarticletitle{The rise of guardians: Fact-checking url
  recommendation to combat fake news}. In \bibinfo{booktitle}{\emph{Proc. of
  SIGIR}}. \bibinfo{pages}{275--284}.
\newblock


\bibitem[Vosoughi et~al\mbox{.}(2018)]%
        {vosoughi2018spread}
\bibfield{author}{\bibinfo{person}{Soroush Vosoughi}, \bibinfo{person}{Deb
  Roy}, {and} \bibinfo{person}{Sinan Aral}.} \bibinfo{year}{2018}\natexlab{}.
\newblock \showarticletitle{The spread of true and false news online}.
\newblock \bibinfo{journal}{\emph{Science}} \bibinfo{volume}{359},
  \bibinfo{number}{6380} (\bibinfo{year}{2018}), \bibinfo{pages}{1146--1151}.
\newblock


\bibitem[Ye and Ling(2019)]%
        {ye2019multi}
\bibfield{author}{\bibinfo{person}{Zhi-Xiu Ye} {and} \bibinfo{person}{Zhen-Hua
  Ling}.} \bibinfo{year}{2019}\natexlab{}.
\newblock \showarticletitle{Multi-Level Matching and Aggregation Network for
  Few-Shot Relation Classification}. In \bibinfo{booktitle}{\emph{Proc. of
  ACL}}. \bibinfo{pages}{2872--2881}.
\newblock


\bibitem[Yin and Roth(2018)]%
        {Yin2018}
\bibfield{author}{\bibinfo{person}{Wenpeng Yin} {and} \bibinfo{person}{Dan
  Roth}.} \bibinfo{year}{2018}\natexlab{}.
\newblock \showarticletitle{TwoWingOS: {A} Two-Wing Optimization Strategy for
  Evidential Claim Verification}. In \bibinfo{booktitle}{\emph{Proc. of
  EMNLP}}, \bibfield{editor}{\bibinfo{person}{Ellen Riloff},
  \bibinfo{person}{David Chiang}, \bibinfo{person}{Julia Hockenmaier}, {and}
  \bibinfo{person}{Jun'ichi Tsujii}} (Eds.). \bibinfo{publisher}{Association
  for Computational Linguistics}, \bibinfo{pages}{105--114}.
\newblock
\urldef\tempurl%
\url{https://doi.org/10.18653/v1/d18-1010}
\showDOI{\tempurl}


\bibitem[Zeng et~al\mbox{.}(2021)]%
        {zeng2021automated}
\bibfield{author}{\bibinfo{person}{Xia Zeng}, \bibinfo{person}{Amani~S
  Abumansour}, {and} \bibinfo{person}{Arkaitz Zubiaga}.}
  \bibinfo{year}{2021}\natexlab{}.
\newblock \showarticletitle{Automated fact-checking: A survey}.
\newblock \bibinfo{journal}{\emph{Language and Linguistics Compass}}
  \bibinfo{volume}{15}, \bibinfo{number}{10} (\bibinfo{year}{2021}),
  \bibinfo{pages}{e12438}.
\newblock


\bibitem[Zhang et~al\mbox{.}(2022)]%
        {zhang2022domain}
\bibfield{author}{\bibinfo{person}{Xin Zhang}, \bibinfo{person}{Yong Jiang},
  \bibinfo{person}{Xiaobin Wang}, \bibinfo{person}{Xuming Hu},
  \bibinfo{person}{Yueheng Sun}, \bibinfo{person}{Pengjun Xie}, {and}
  \bibinfo{person}{Meishan Zhang}.} \bibinfo{year}{2022}\natexlab{}.
\newblock \showarticletitle{Domain-Specific NER via Retrieving Correlated
  Samples}. In \bibinfo{booktitle}{\emph{Proc. of COLING}}.
  \bibinfo{pages}{2398--2404}.
\newblock


\end{thebibliography}










\end{document}